\documentclass{article}
\usepackage{ijcai25}

\usepackage{times}
\usepackage{soul}
\usepackage{url}
\usepackage[hidelinks]{hyperref}
\usepackage[utf8]{inputenc}
\usepackage[small]{caption}
\usepackage{graphicx}
\graphicspath{ {./figures/} }
\usepackage{amsmath}
\usepackage{amsthm}
\usepackage{booktabs}
\usepackage{algorithm}
\usepackage{algorithmic}
\usepackage[switch]{lineno}
\usepackage[table]{xcolor}
\usepackage{natbib}
\usepackage{dsfont}

\title{Wisdom from Diversity: Bias Mitigation Through Hybrid Human-LLM Crowds}

\author{
Axel Abels$^{1,2,3,}$\footnote{Corresponding author. Email: axel.abels@ulb.be}
\and
Tom Lenaerts$^{1,2,3,4}$\\
\affiliations
$^1$Machine Learning Group, Université Libre de Bruxelles\\
$^2$AI Lab, Vrije Universiteit Brussel\\
$^3$FARI, AI for the Common-Good Institute, ULB-VUB\\
$^4$Center for Human-Compatible AI, UC Berkeley\\
}

\begin{document}
\maketitle

{
\renewcommand\thefootnote{}%
\footnotetext{\hspace*{-0em}Accepted for publication in the Proceedings of the 34th International Joint Conference on Artificial Intelligence (IJCAI 2025).}
% \addtocounter{footnote}{-1} % keep numbering consistent
}
\begin{abstract}
Despite their performance, large language models (LLMs) can inadvertently perpetuate biases found in the data they are trained on. By analyzing LLM responses to bias-eliciting headlines, we find that these models often mirror human biases. To address this, we explore crowd-based strategies for mitigating bias through response aggregation. We first demonstrate that simply averaging responses from multiple LLMs, intended to leverage the ``wisdom of the crowd", can exacerbate existing biases due to the limited diversity within LLM crowds. In contrast, we show that locally weighted aggregation methods more effectively leverage the wisdom of the LLM crowd, achieving both bias mitigation and improved accuracy. Finally, recognizing the complementary strengths of LLMs (accuracy) and humans (diversity), we demonstrate that hybrid crowds containing both significantly enhance performance and further reduce biases across ethnic and gender-related contexts.
\end{abstract}

\section{Introduction}

The increasing adoption of LLM assistants raises concerns about their potential to perpetuate or amplify societal biases \citep{bolukbasi2016man,caliskan2017semantics, zhao2017men, sheng2019woman, zhang2020demographics}.  These often subtle yet pervasive stereotypes pose a significant challenge to their responsible use.  While prior work has identified biases stemming from training data and model architectures \citep{blodgett2020language,bender2021dangers}, understanding how these biases manifest---especially in comparison to human biases---remains crucial. This understanding is key not only for evaluating LLM fairness and reliability but also for designing systems that complement human decision-making without perpetuating harmful biases.

In this work, we analyze LLM responses to a set of bias-eliciting headlines, comparing them to previously collected human responses \citep{abels2024mitigating}. This allows us to assess how LLM biases align or differ from human biases, as well as to evaluate bias patterns across various LLMs.

Building on these insights, we explore strategies to mitigate biases in LLM outputs. Unlike previous works which tackled biases of individuals LLMs \citep{zhang2020demographics, zhao2018gender, tamkin2023evaluating}, we draw on the principles of collective intelligence \citep{surowiecki2005wisdom,malone2022handbook}. Specifically, we investigate the effectiveness of aggregating responses from multiple LLMs---creating ``LLM crowds". We hypothesize that, similar to human crowds \citep{abels2024mitigating}, this aggregation can effectively diminish bias by leveraging the diversity of responses. Using methods like simple averaging and locally weighted averages---where weights are tailored to the specializations of the crowd---we evaluate how effectively these approaches address biases while enhancing performance.

Finally, given the complementary strengths of humans (high diversity, lower individual accuracy) and LLMs (high individual accuracy, lower diversity), and recognizing that both are crucial for effective collective intelligence \citep{hong2004groups, surowiecki2005wisdom}, we examine the potential of hybrid crowds. 
Our findings demonstrate that hybrid crowds significantly improve accuracy while reducing biases to negligible levels, outperforming both LLM-only and human-only groups. These results highlight the potential of hybrid approaches to enhance fairness and accuracy in applications such as content moderation and hiring systems.

To summarize, our contributions include: (1) a comparative analysis of LLM and human biases on bias-eliciting headlines, (2) an evaluation of the potential and limitations of LLM crowds, and (3) a demonstration of the superior performance and fairness of hybrid crowds.

\section{Background}\label{sec:background}
Social biases are deeply ingrained cognitive patterns that influence human perception, judgment, and behavior. These biases, often unconscious, arise from stereotypes, cultural norms, and societal structures. While they are often the product of heuristics---mental shortcuts which enable quick decision-making---they frequently result in discrimination, inequality, and harm to marginalized groups \citep{greenwald1998measuring, devine1989stereotypes}. For instance, racial biases perpetuate systemic inequalities in education, employment, and healthcare \citep{williams1996racism,reskin2000proximate}, while gender biases limit opportunities and reinforce harmful stereotypes about abilities and roles \citep{eagly1990gender}. 

Despite growing awareness and efforts to combat them, social biases persist, influencing hiring practices, educational access, healthcare delivery, and legal outcomes \citep{bertrand2004emily, pager2008marked}.  These biases are often perpetuated through societal institutions and media representations \citep{dovidio2017reducing}.
The growing integration of artificial intelligence into daily life adds a new dimension to this challenge. Widely adopted LLM assistants could expose users to these biases on a broader scale, reflecting and potentially reinforcing prejudices embedded in their training data.

\subsection{Social Biases in LLMs}
LLMs are a class of neural networks trained on extensive text corpora to perform a wide range of natural language processing (NLP) tasks, including text generation, summarization, translation, and conversational interactions \citep{vaswani2017attention, radford2019language, brown2020language}. Building on earlier NLP advancements, such as word embeddings \citep{mikolov2013efficient}, LLMs use transformer-based architectures to create dynamic, context-aware representations \citep{vaswani2017attention}, enabling them to achieve high accuracy and fluency in complex tasks. 
Their versatility and human-like responses have made them integral to applications such as virtual assistants and content creation \citep{bommasani2021opportunities}. However, their reliance on vast amounts of real-world data also makes them susceptible to inheriting and, in some cases, amplifying societal biases present in their training data \citep{bolukbasi2016man, bender2021dangers, sheng2019woman}.

These biases manifest in various forms, including biased word associations \citep{caliskan2017semantics}, stereotypical sentence generation \citep{zhao2017men}, and unequal performance across demographic groups \citep{zhang2020demographics}. This phenomenon has been extensively documented across various social dimensions, including gender, race, and socioeconomic status \citep{bolukbasi2016man, davidson2019racial, caliskan2017semantics}. For instance, word embedding association tests reveal that LLMs often replicate harmful stereotypes, such as associating certain professions predominantly with one gender \citep{zhao2018gender}. Such biases extend to conversational systems, which risk  exposing users to outputs that reinforce stereotypes and perpetuate societal prejudices. 

Understanding how LLMs manifest social biases has therefore become a crucial component in the development of socially responsible AI systems \citep{team2023gemini,TheC3,achiam2023gpt}. However, mitigating these biases is a complex challenge. Early efforts focused on post-processing techniques, such as neutralizing gender-specific dimensions in word embeddings \citep{bolukbasi2016man}. More recent research includes interventions during the generation phase, such as reweighting or filtering outputs to suppress stereotypes \citep{sheng2020towards, liang2021towards}.

Beyond algorithmic solutions, socio-technical approaches have gained attention as complementary strategies. These include curating representative datasets, incorporating diverse evaluation metrics, and prioritizing interdisciplinary collaboration to address ethical concerns \citep{birhane2021multimodal, bender2021dangers}. Such approaches aim to tackle biases at their source, emphasizing the importance of combining technical innovations with ethical and societal considerations. 

Despite these efforts, several challenges persist. Bias mitigation techniques can struggle to generalize across different bias types, and often degrade overall model performance. Additionally, the high cost of training state-of-the-art models has incentivized the development of closed-source LLMs, complicating efforts to ensure transparency and accountability.

\subsection{Collective Intelligence}
Beyond model-centric solutions, we argue in this work that leveraging collective intelligence offers a promising complementary approach to bias mitigation. The principle that groups can achieve more accurate and reliable outcomes than individuals is well-established \citep{condorcet1785essay, hong2004groups, surowiecki2005wisdom}. This ``wisdom of the crowd" effect leverages diverse perspectives and independent errors to improve decision-making.  

\paragraph{Wisdom of LLM Crowds}
The concept of crowd wisdom, widely studied in human groups and exploited by traditional machine learning ensembles \citep{surowiecki2005wisdom,malone2022handbook,wolpert1992stacked,breiman1996stacked}, is now being explored in the context of LLMs. Recent research demonstrates the existence of a ``wisdom of LLM crowds". For instance, \cite{schoenegger2024wisdom} show that LLM ensembles achieve forecasting accuracy comparable to human crowds. Additionally, they show that exposing LLMs to the median prediction of human crowds enhances their accuracy, consistent with findings from human studies \citep{becker2017network}.

Exploring bias dynamics, \cite{chuang2024wisdom} show that LLM crowds, when simulating partisan personas, can mimic human-like partisan biases. However, deliberation among these personas leads to more accurate beliefs, mirroring the benefits of discussion and information sharing which are sometimes observed in human groups \citep{becker2019wisdom}.

While these studies reproduce key results in LLMs, they leave open critical questions. Specifically, how should LLMs be aggregated to systematically mitigate biases while enhancing performance? Furthermore, the potential synergy between LLMs and humans remains largely unexplored. To enable us to address these gaps, we now review prior methods to mitigate biases and restore crowd wisdom in human groups.

\subsection{Bias Mitigation in Human Crowds}

Understanding how biases influence crowd wisdom in human groups provides valuable insights for developing aggregation strategies in LLM and hybrid crowds. The following experiment illustrates the prevalence of social biases in individuals and their impact on collective decision-making.

\subsubsection{The Headline Experiment}

In \citep{abels2024mitigating}, participants evaluated a balanced set of genuine and altered news headlines, where demographic groups were swapped to create counterfactual pairs. For example, the headline \textit{``Men more likely than women to say they are financially better off since last year"} was altered to \textit{``Women more likely than men to say they are financially better off since last year"}. Headlines described positive or negative outcomes for various demographic groups (gender, ethnicity, age), and participants rated their authenticity on a scale from ``very unlikely" to ``very likely."

This design allowed the authors to measure bias by comparing error rates across demographic groups and outcomes. For example, discrepancies in error rates for positive vs. negative outcomes for white individuals revealed underlying biases. Further analyses of the responses revealed that factors like responders' demographics, headline categories, and question framing significantly influenced individual judgments, often leading to systematic errors. 

\subsubsection{Restoring Crowd Wisdom in Human Groups}
While the ``wisdom of the crowd" suggests that diverse perspectives could mitigate individual biases, \cite{abels2024mitigating} demonstrated that dominant societal beliefs can still lead to collective errors. To address this, they proposed using locally-weighted aggregation to counter individual biases and variations in expertise. 
Specifically, they used ExpertiseTrees \citep{pmlr-v202-abels23a}, which leverage the diversity within a group by partitioning the context space (e.g., headline categories) and fitting specialized models to each region. Similar to decision trees, ExpertiseTrees dynamically introduce splits only when they improve group aggregation. 

This hierarchical approach offers several advantages compared to other aggregation methods. While techniques like stacking \citep{breiman1996stacked} or simple averaging provide static aggregation, ExpertiseTrees allow for more nuanced aggregation by adjusting individuals' weights to the problem (here, the headline) at hand. This can be particularly beneficial for bias mitigation, as different individuals may exhibit different biases in different contexts. In particular, \cite{abels2024mitigating} show that when identifying fake news, humans display performance that varies significantly with the demographic group contained in the headline. ExpertiseTrees partition the context space to route inputs to individuals most likely to provide unbiased responses, improving accuracy and fairness. %This adaptability  results in improved collective intelligence in human crowds \citep{abels2024mitigating}.

While this research highlights the promise of advanced aggregation methods in human crowds, their applicability to LLM crowds remains uncertain. LLMs may lack the diversity found in humans and exhibit distinct biases, potentially limiting the effectiveness of aggregation strategies. Additionally, we hypothesize that the complementary strengths of humans and LLMs---humans contributing greater diversity and LLMs offering higher individual accuracy---present a unique opportunity for hybrid approaches. With these considerations in mind, we now present how we addressed these challenges.

\section{Methods}\label{sec:methods}

Our aim in this work is two-fold. First, we aim to compare LLM biases with those of humans. Second, we aim to study the benefits offered by crowds, first crowds of LLMs, and then hybrid crowds, containing both LLMs and humans. As a first step, we replicate the headline experiment conducted by \cite{abels2024mitigating} on LLMs, enabling a direct comparison of LLM biases and performance with those of humans.

To ensure broad coverage, we include both closed-source and open-source LLMs from OpenAI \citep{achiam2023gpt}, Anthropic \citep{TheC3}, Google \citep{team2023gemini}, Meta \citep{dubey2024llama}, Mistral \citep{mistral2023chat}, Alibaba \citep{bai2023qwen}, and DeepSeek AI \citep{liu2024deepseek}. %These models power widely available chat assistants such as ChatGPT \citep{openai2022chatgpt}, Google Gemini \citep{google2023gemini}, and Mistral Chat \citep{mistral2023chat}. 
Supplementary Table S.2 provides a detailed comparison of the selected models. These LLMs were selected to capture a diverse range of architectures and training paradigms, ensuring a comprehensive analysis of LLM performance and biases.

\subsection{Experimental Procedure}\label{sec:experiments}
We closely replicate the methodology of \citep{abels2024mitigating} to enable a direct comparison between humans and LLMs.

We therefore prompted LLMs to estimate the likelihood that the given headlines were real. Following the human study, LLMs were instructed to respond using a 5-point Likert scale ranging from ``very unlikely" to ``very likely." The response for every headline $h$ was then mapped onto a numerical likelihood: $p_h \in \{0, 0.25, 0.5, 0.75, 1\}$. All LLMs were prompted in a 4-shot setting with instructions designed to replicate the guidance provided to human participants. Prompts included a brief explanation of the task and example responses. Details on the prompting procedure and parameters are given in the supplementary materials.

\subsection{Metrics}\label{sec:metrics}

We evaluate LLMs on their accuracy, diversity, and susceptibility to biases, comparing them with previously collected human data. 

\paragraph{Accuracy}
Accuracy is calculated as the proportion of correctly identified headlines, whether genuine or altered. Since the dataset is balanced across class (genuine vs. altered), sentiment (positive vs. negative), and demographic groups (man $\leftrightarrow$ woman, young $\leftrightarrow$ old, White $\leftrightarrow$ African American), this metric provides an overall indication of performance. To identify potential specializations, accuracy is also reported for each combination of class and demographic group.

To further identify whether there are systematic differences in accuracy which can be attributed to prejudices against certain groups, we explore the following types of bias. 

\paragraph{Counterfactual Bias}
In line with conditional statistical parity \citep{corbett2017algorithmic}, we define counterfactual bias as a systematic tendency to favor certain outcomes for specific demographic groups. For a demographic group $g$ and its counterpart $g'$, this bias is quantified as:
\begin{equation}
\Delta_{g,g'}(s, \sigma)
\;=\;
\mathds{E}_{h \sim H_{s,\sigma,g}}[p_h]
\;-\;
\mathds{E}_{h' \sim H_{s,\sigma,g'}}[p_{h'}],
\label{eq:groupbias}
\end{equation}
where $p_h$ is the likelihood assigned to headline $h$, and $H_{s,\sigma,g}$ is the subset of headlines of status $s$ (Genuine or Altered), sentiment $\sigma$  (positive or negative), and demographic group $g$ (e.g., man, woman, young , old, White, African American).

Positive values indicate a bias towards group $g$, whereas negative values indicate a bias towards $g'$.  For example, if a model is more likely to believe headlines reporting positive outcomes for men over women, $\Delta_{man,woman}(genuine, $+$)$ will be positive, indicating a bias in favor of men. In practice, $\mathds{E}_{h \sim H_{s,\sigma,g}}[p_h]$ is estimated as the mean likelihood assigned to all headlines in $H_{s,\sigma,g}$. 
To determine the significance of counterfactual biases, we use Mann-Whitney U tests \citep{mann1947test} to compare the distributions of likelihoods assigned to headlines from different demographic groups. 

\paragraph{Framing Effects}
Framing effects occur when responses differ based on how identical information is presented. In the headline dataset, a responder's likelihood for a headline $h$ may differ from $1-p_{h'}$, where $h'$ is the counterfactual variant of $h$ obtained by swapping demographic groups. Systematic differences indicate susceptibility to framing effects.

%To measure framing effects in the headline task, we analyze pairs of genuine-altered headlines. Significant framing effects occur if responses to these pairs differ systematically.
For a group $g$ and sentiment $\sigma$, average framing effects are: 
$$ \Delta_F(\sigma,g) = \frac{1}{|H_{\sigma,g}|}\sum_{h\in H_{\sigma,g}} p_h - (1-p_{h'}) $$
where $H_{\sigma,g}$ is the set of headlines with that sentiment and group. For example, for humans both $\Delta_F(positive,young)$ and  $\Delta_F(negative,young)$ are positive, indicating they are likely to assign the same belief to a headline reporting the opposite outcome for age groups.

To assess significance, we use Wilcoxon signed-rank tests \citep{wilcoxon1992individual} on the set of differences.

\paragraph{Diversity}
Diversity is crucial for crowd wisdom, as it leverages individuals' independent errors \citep{wood2023unified}. When group members consistently make the same mistakes, their combination offers little benefit, as they merely amplify their shared errors and biases. Conversely, when individuals exhibit diverse behavior---where one member's errors are offset by other members' correct predictions---the group can achieve higher performance. This is particularly relevant for bias mitigation, as one group member might exhibit gender bias in a specific context, while another model, trained on a different dataset, or using a different architecture, might not.

To quantify diversity, we use the Q-statistic \citep{yule1900vii}, which measures the extent to which two classifiers make the same predictions (correct or incorrect). It is defined as %Formally, for two classifiers $C_1$ and $C_2$, the Q-statistic is defined as:
\begin{equation}
Q \;=\; \frac{N_{11}\,N_{00} - N_{10}\,N_{01}}{N_{11}\,N_{00} + N_{10}\,N_{01}},\label{eq:qstatistic}
\end{equation}
where $N_{11}$ is the number of instances both classifiers predict correctly, $N_{00}$ is the number of instances both classifiers predict incorrectly,  $N_{10}$ is the number of instances in which $C_1$ is correct while $C_2$ is incorrect, and $N_{01}$ is the number of instances in which $C_1$ is incorrect while $C_2$ is correct. 

%A $Q$-statistic close to $+1$ indicates strong agreement, implying low diversity (i.e., both classifiers tend to misclassify the same headlines). A value near $0$ suggests near-independence, while negative values indicate a systematic tendency to disagree, indicating high diversity. 

Combined with high individual accuracies, a lower $Q$-statistic often signals useful diversity, yielding better ensembles as the collective can compensate for individual mistakes.

\subsection{Ensembling Strategies}\label{sec:ensembling}

Our second major contribution is the exploration of ensemble-based strategies to mitigate LLM biases. We compare LLM crowds against individual models to highlight the benefits offered by different aggregation strategies, including potential trade-offs between accuracy and bias mitigation.

In addition to using simple averages, we investigate two types of weighted averages. First, we explore traditional stacking \citep{breiman1996stacked}, wherein group members are assigned a constant weight. While such weights effectively prioritize consistently well-performing members, they cannot adapt to variations in context, such as biases or specializations specific to certain headline categories. To address this limitation, we also study weights tailored to headline categories. 

Specifically, let $p^m_h$ be the likelihood group member $m$ assigns to headline $h$. The aggregated likelihood is:
\[
\bar{p}_h \;=\; \sum_m \, w^m_{\phi(h)} \, p^m_h,
\]
where $\phi(h)$ is the headline category (age, gender, ethnicity), and $w^m_{\phi(h)}$ is the weight of individual $m$ for that category. 

We use ExpertiseTrees (see \citep{pmlr-v202-abels23a}, as well as our detailed description in the Supplementary Information) to learn these localized weights\footnote{For both weighted average methods, we use cross-validation to ensure weights were not trained on headlines they are evaluated on.}. Unlike traditional stacking, which learns static weights for combining model predictions, ExpertiseTrees use a tree structure to assign weights based on input-specific characteristics, such as headline categories. This allows ExpertiseTrees to amplify the contributions of less biased members while down-weighting biased predictions, thereby improving fairness and accuracy. In the headline setting, localized weights allow us to adapt the aggregation to headline categories.  For instance, \cite{abels2024mitigating} demonstrate that human performance varies significantly across headline categories. ExpertiseTrees exploit this specialization by learning a distinct set of weights for each category, allowing for context-sensitive aggregation.

\paragraph{Benchmark-based Sampling} When forming LLM groups, we employ two sampling approaches. First, we use random sampling, where LLMs are selected uniformly at random from the pool of available models. This serves as a baseline for evaluating the general benefits of ensemble methods. Second, we select LLMs based on their scores on the widely used MMLU benchmark \citep{hendrycks2020measuring}. This method prioritizes high-performing models, which we hypothesize will better correlate with improved performance on the headline task. However, selecting high-performing models may reduce diversity due to increased similarity among the selected models, presenting a trade-off between individual strength and collective diversity. Groups formed using this performance-based approach are denoted with a ``+" sign, e.g., LLM+. 

To evaluate the potential of hybrid ensembles, we combined human responders from the headline experiment dataset with LLMs. We hypothesize that hybrid groups leverage the complementary strengths of humans and LLMs: humans provide greater diversity, while LLMs offer higher individual accuracy. The LLMs in hybrid groups were selected using the same two sampling approaches, resulting in two types of hybrid groups: hybrid, with randomly sampled LLMs, and hybrid+, with performance-selected LLMs.

To assess the impact of group size on collective performance, we evaluated groups ranging in size from 2 to 16 responders. This range reflects practical limits, as the available pool of LLMs consisted of 18 models. Although larger groups were not tested, observed trends provide a basis for inferring potential outcomes for larger groups.

\section{Results}\label{sec:results}

We now present the results of our evaluation of the headline experiment on LLMs\footnote{Code to reproduce these results is available at \url{https://github.com/axelabels/HybridCrowds}}. Specifically, we first compare the performance and bias of individual LLMs to human participants. Next, we demonstrate the limitations of LLM crowds and the advantages of hybrid crowds, emphasizing the potential of locally weighted aggregations in achieving improved outcomes.

\begin{table*}[ht]
\centering
\resizebox{.87\textwidth}{!}{%
\begin{tabular}{l cc cc cc r}
\toprule
 & \multicolumn{2}{c}{A\textsc{ge}} & \multicolumn{2}{c}{E\textsc{thnicity}} & \multicolumn{2}{c}{G\textsc{ender}} & A\textsc{verage} \\ 
M\textsc{odel} & altered & genuine & altered & genuine & altered & genuine \\ \midrule
human &  0.44 ($\Delta$=+0.01) &  0.65 ($\Delta$=-0.01) &  0.65 ($\Delta$=+0.09) & \cellcolor[HTML]{FFCCCC} 0.48 ($\Delta$=+0.17) &  0.57 ($\Delta$=-0.07) &  0.53 ($\Delta$=-0.04) & 0.550 \\
 \midrule
Qwen2.5-72B-Instruct &  0.57 ($\Delta$=+0.04) &  0.82 ($\Delta$=-0.05) &  0.65 ($\Delta$=+0.12) & \cellcolor[HTML]{FFCCCC} 0.62 ($\Delta$=+0.14) &  0.65 ($\Delta$=+0.04) &  0.50 ($\Delta$=+0.04) & 0.637 \\
claude-3-5-haiku &  0.78 ($\Delta$=+0.03) &  0.35 ($\Delta$=+0.09) &  0.52 ($\Delta$=+0.16) & \cellcolor[HTML]{FF6666} 0.47 ($\Delta$=+0.33) &  0.80 ($\Delta$=+0.00) &  0.52 ($\Delta$=-0.12) & 0.575 \\
claude-3-5-sonnet &  0.55 ($\Delta$=+0.06) &  0.88 ($\Delta$=+0.11) & \cellcolor[HTML]{FF6666} 0.55 ($\Delta$=+0.35) & \cellcolor[HTML]{FF6666} 0.75 ($\Delta$=+0.32) &  0.68 ($\Delta$=-0.11) &  0.82 ($\Delta$=-0.03) & 0.704 \\
claude-3-opus &  0.50 ($\Delta$=+0.10) &  0.93 ($\Delta$=+0.08) & \cellcolor[HTML]{FFCCCC} 0.48 ($\Delta$=+0.18) &  0.73 ($\Delta$=+0.06) & \cellcolor[HTML]{FFCCCC} 0.65 ($\Delta$=-0.15) &  0.62 ($\Delta$=+0.01) & 0.650 \\
deepseek &  0.40 ($\Delta$=-0.03) &  0.85 ($\Delta$=+0.08) & \cellcolor[HTML]{FF6666} 0.42 ($\Delta$=+0.23) & \cellcolor[HTML]{FF6666} 0.65 ($\Delta$=+0.27) & \cellcolor[HTML]{FF9999} 0.45 ($\Delta$=-0.17) &  0.75 ($\Delta$=-0.01) & 0.588 \\
gemini-1.5-flash &  0.48 ($\Delta$=+0.03) &  0.60 ($\Delta$=+0.00) & \cellcolor[HTML]{FF9999} 0.68 ($\Delta$=+0.22) & \cellcolor[HTML]{FF6666} 0.50 ($\Delta$=+0.28) &  0.78 ($\Delta$=-0.02) &  0.42 ($\Delta$=-0.02) & 0.575 \\
gemini-1.5-pro &  0.62 ($\Delta$=+0.06) &  0.53 ($\Delta$=-0.01) & \cellcolor[HTML]{FF6666} 0.62 ($\Delta$=+0.24) & \cellcolor[HTML]{FF6666} 0.68 ($\Delta$=+0.24) & \cellcolor[HTML]{FFCCCC} 0.75 ($\Delta$=-0.14) &  0.45 ($\Delta$=-0.05) & 0.608 \\
gemini-2.0-flash &  0.45 ($\Delta$=+0.04) &  0.77 ($\Delta$=+0.08) & \cellcolor[HTML]{FFCCCC} 0.45 ($\Delta$=+0.19) & \cellcolor[HTML]{FF6666} 0.70 ($\Delta$=+0.32) & \cellcolor[HTML]{FFCCCC} 0.60 ($\Delta$=-0.16) & \cellcolor[HTML]{FFCCCC} 0.70 ($\Delta$=-0.15) & 0.612 \\
gemma2-9b &  0.33 ($\Delta$=+0.01) &  0.78 ($\Delta$=-0.01) & \cellcolor[HTML]{FFCCCC} 0.47 ($\Delta$=+0.16) & \cellcolor[HTML]{FF6666} 0.62 ($\Delta$=+0.27) &  0.57 ($\Delta$=-0.10) &  0.65 ($\Delta$=+0.06) & 0.571 \\
gemma-2-27b &  0.53 ($\Delta$=+0.02) &  0.62 ($\Delta$=-0.01) &  0.80 ($\Delta$=+0.01) & \cellcolor[HTML]{FF6666} 0.38 ($\Delta$=+0.27) & \cellcolor[HTML]{FF9999} 0.82 ($\Delta$=-0.14) &  0.40 ($\Delta$=-0.11) & 0.592 \\
gpt-4 &  0.65 ($\Delta$=-0.01) &  0.57 ($\Delta$=-0.05) &  0.70 ($\Delta$=+0.06) & \cellcolor[HTML]{FF6666} 0.45 ($\Delta$=+0.31) &  0.70 ($\Delta$=-0.14) &  0.50 ($\Delta$=-0.02) & 0.596 \\
gpt-4-turbo &  0.55 ($\Delta$=-0.01) &  0.68 ($\Delta$=-0.01) &  0.65 ($\Delta$=+0.03) & \cellcolor[HTML]{FF9999} 0.55 ($\Delta$=+0.22) &  0.75 ($\Delta$=-0.10) &  0.60 ($\Delta$=+0.04) & 0.629 \\
gpt-4o &  0.57 ($\Delta$=-0.08) &  0.80 ($\Delta$=-0.04) & \cellcolor[HTML]{FF9999} 0.68 ($\Delta$=+0.18) & \cellcolor[HTML]{FF6666} 0.75 ($\Delta$=+0.24) &  0.78 ($\Delta$=-0.08) &  0.75 ($\Delta$=-0.05) & \textbf{0.721} \\
gpt-4o-mini &  0.28 ($\Delta$=-0.06) &  0.82 ($\Delta$=+0.01) &  0.62 ($\Delta$=+0.08) & \cellcolor[HTML]{FF6666} 0.50 ($\Delta$=+0.29) & \cellcolor[HTML]{FF9999} 0.42 ($\Delta$=-0.16) & \cellcolor[HTML]{FFCCCC} 0.72 ($\Delta$=-0.11) & 0.562 \\
Llama-3.3-70B &  0.43 ($\Delta$=+0.13) &  0.80 ($\Delta$=+0.10) & \cellcolor[HTML]{FF9999} 0.33 ($\Delta$=+0.20) & \cellcolor[HTML]{FF9999} 0.80 ($\Delta$=+0.17) &  0.53 ($\Delta$=-0.13) &  0.65 ($\Delta$=-0.10) & 0.588 \\
mistral-large &  0.50 ($\Delta$=+0.04) &  0.80 ($\Delta$=+0.05) & \cellcolor[HTML]{FF9999} 0.62 ($\Delta$=+0.20) & \cellcolor[HTML]{FF6666} 0.57 ($\Delta$=+0.33) &  0.78 ($\Delta$=-0.10) &  0.68 ($\Delta$=-0.07) & 0.658 \\
mixtral-8x7b &  0.28 ($\Delta$=+0.06) &  0.90 ($\Delta$=+0.03) &  0.28 ($\Delta$=+0.04) & \cellcolor[HTML]{FF9999} 0.80 ($\Delta$=+0.19) & \cellcolor[HTML]{FF9999} 0.35 ($\Delta$=-0.15) &  0.88 ($\Delta$=-0.01) & 0.579 \\
open-mistral &  0.73 ($\Delta$=-0.07) &  0.38 ($\Delta$=+0.05) &  0.93 ($\Delta$=-0.06) & \cellcolor[HTML]{FFCCCC} 0.25 ($\Delta$=+0.15) & \cellcolor[HTML]{FFCCCC} 0.80 ($\Delta$=-0.14) &  0.25 ($\Delta$=+0.02) & 0.554 \\
\midrule
average(LLM) &  0.53 ($\Delta$=+0.02) &  0.83 ($\Delta$=+0.03) & \cellcolor[HTML]{FF9999} 0.57 ($\Delta$=+0.14) & \cellcolor[HTML]{FF6666} 0.62 ($\Delta$=+0.24) & \cellcolor[HTML]{FFCCCC} 0.72 ($\Delta$=-0.11) &  0.65 ($\Delta$=-0.04) & 0.652 \\
average(human) &  0.43 ($\Delta$=+0.01) &  0.75 ($\Delta$=-0.02) & \cellcolor[HTML]{FF6666} 0.79 ($\Delta$=+0.11) & \cellcolor[HTML]{FF6666} 0.48 ($\Delta$=+0.17) & \cellcolor[HTML]{FFCCCC} 0.67 ($\Delta$=-0.09) &  0.55 ($\Delta$=-0.04) & 0.611 \\
average(hybrid) &  0.50 ($\Delta$=+0.01) &  0.80 ($\Delta$=+0.00) & \cellcolor[HTML]{FF9999} 0.69 ($\Delta$=+0.13) & \cellcolor[HTML]{FF6666} 0.56 ($\Delta$=+0.21) & \cellcolor[HTML]{FF9999} 0.69 ($\Delta$=-0.10) &  0.60 ($\Delta$=-0.04) & 0.641 \\
average(LLM+) &  0.60 ($\Delta$=+0.01) &  0.83 ($\Delta$=+0.03) & \cellcolor[HTML]{FF9999} 0.52 ($\Delta$=+0.17) & \cellcolor[HTML]{FF6666} 0.62 ($\Delta$=+0.24) & \cellcolor[HTML]{FFCCCC} 0.70 ($\Delta$=-0.11) &  0.65 ($\Delta$=-0.02) & 0.654 \\
average(hybrid+) &  0.55 ($\Delta$=+0.01) &  0.78 ($\Delta$=+0.02) & \cellcolor[HTML]{FF6666} 0.73 ($\Delta$=+0.18) & \cellcolor[HTML]{FF6666} 0.60 ($\Delta$=+0.23) & \cellcolor[HTML]{FF9999} 0.77 ($\Delta$=-0.11) &  0.63 ($\Delta$=-0.05) & \textbf{0.678} \\
\midrule
WeightedAverage(LLM) &  0.67 ($\Delta$=-0.01) &  0.68 ($\Delta$=-0.02) &  0.67 ($\Delta$=+0.11) & \cellcolor[HTML]{FF9999} 0.70 ($\Delta$=+0.14) &  0.74 ($\Delta$=-0.04) &  0.75 ($\Delta$=+0.01) & 0.703 \\
WeightedAverage(human) &  0.62 ($\Delta$=-0.02) &  0.66 ($\Delta$=-0.04) &  0.69 ($\Delta$=+0.08) & \cellcolor[HTML]{FFCCCC} 0.70 ($\Delta$=+0.12) &  0.70 ($\Delta$=-0.09) &  0.65 ($\Delta$=-0.04) & 0.671 \\
WeightedAverage(hybrid) &  0.66 ($\Delta$=-0.01) &  0.68 ($\Delta$=-0.02) &  0.69 ($\Delta$=+0.07) & \cellcolor[HTML]{FF9999} 0.71 ($\Delta$=+0.11) &  0.75 ($\Delta$=-0.07) &  0.74 ($\Delta$=-0.03) & 0.704 \\
WeightedAverage(LLM+)  &  0.76 ($\Delta$=-0.02) &  0.75 ($\Delta$=-0.03) & \cellcolor[HTML]{FFCCCC} 0.69 ($\Delta$=+0.12) & \cellcolor[HTML]{FF9999} 0.72 ($\Delta$=+0.15) &  0.72 ($\Delta$=-0.02) &  0.79 ($\Delta$=+0.01) & 0.738 \\
WeightedAverage(hybrid+) &  0.71 ($\Delta$=-0.02) &  0.74 ($\Delta$=+0.00) & \cellcolor[HTML]{FF9999} 0.75 ($\Delta$=+0.11) & \cellcolor[HTML]{FF9999} 0.77 ($\Delta$=+0.13) &  0.83 ($\Delta$=-0.06) &  0.83 ($\Delta$=-0.03) & \textbf{0.772} \\
\midrule
ExpertiseTree(LLM) &  0.70 ($\Delta$=-0.02) &  0.70 ($\Delta$=-0.03) &  0.64 ($\Delta$=+0.07) & \cellcolor[HTML]{FF9999} 0.65 ($\Delta$=+0.14) &  0.76 ($\Delta$=-0.03) &  0.75 ($\Delta$=-0.04) & 0.701 \\
ExpertiseTree(human) &  0.72 ($\Delta$=-0.02) &  0.76 ($\Delta$=-0.02) &  0.75 ($\Delta$=+0.05) &  0.77 ($\Delta$=+0.08) &  0.75 ($\Delta$=-0.04) &  0.75 ($\Delta$=-0.02) & 0.749 \\
ExpertiseTree(hybrid) &  0.73 ($\Delta$=+0.01) &  0.80 ($\Delta$=-0.01) &  0.73 ($\Delta$=+0.05) &  0.76 ($\Delta$=+0.05) &  0.82 ($\Delta$=-0.04) &  0.83 ($\Delta$=-0.03) & 0.777 \\
ExpertiseTree(LLM+) &  0.74 ($\Delta$=-0.01) &  0.75 ($\Delta$=-0.03) &  0.68 ($\Delta$=+0.09) &  0.71 ($\Delta$=+0.11) &  0.82 ($\Delta$=-0.01) &  0.80 ($\Delta$=-0.01) & 0.752 \\
ExpertiseTree(hybrid+) &  0.81 ($\Delta$=-0.02) &  0.82 ($\Delta$=-0.03) &  0.75 ($\Delta$=+0.04) &  0.75 ($\Delta$=+0.04) &  0.89 ($\Delta$=-0.00) &  0.86 ($\Delta$=-0.03) & \textbf{0.813} \\
\bottomrule
\end{tabular}}
\caption{Accuracy and counterfactual bias ($\Delta$, see Equation \ref{eq:groupbias}) across headline categories. High counterfactual biases indicate a higher belief in positive headlines for historically privileged groups (older, white, male). Cell shading represents statistical significance of the counterfactual bias: darkest red for $p<0.01$, medium red for $p<0.05$, and light red for $p<0.1$. Rows labeled  \textit{average($\cdot$)},\textit{WeightedAverage($\cdot$)}, and \textit{ExpertiseTree($\cdot$)} give the performance of groups of 8 aggregated through respectively simple averages, weighted averages, or locally weighted averages. These groups can consist of humans, LLMs, or a mix of both (hybrid). Instead of randomly sampling available LLMs, LLM+ and Hybrid+ select the models with the highest scores on the MMLU benchmark.  }\label{tab:performance}
\end{table*}

\subsection{LLM Performance}
To evaluate the potential for collective intelligence, we begin by analyzing individual LLM performances. Table \ref{tab:performance} reports the accuracy of various LLMs across headline categories, along with their counterfactual biases. For comparison, the table also includes the average human participant.

\paragraph{LLMs show above-human-level accuracy}
Our findings indicate that GPT-4o and Claude-3.5-Sonnet achieve the highest average accuracy, consistent with prior benchmarks like MMLU \citep{hendrycks2020measuring}. While smaller models, such as the open-source mistral-8x7b, generally perform worse, all tested LLMs outperform the average human on this task.

\paragraph{LLMs mirror human counterfactual biases}
Similar to human responders, all tested LLMs show some degree of counterfactual bias, especially for headlines involving ethnic groups. LLMs tend to assign higher likelihoods to headlines reporting positive outcomes for White individuals than for African-American ones, especially when only considering genuine headlines (Table \ref{tab:performance}, \textsc{Ethnicity}-Genuine column). All but one LLM show a significant effect for this bias. 

For age headlines, counterfactual bias is less pronounced in both LLMs and humans. Gender headlines elicit moderate bias, with fewer than half of the tested LLMs showing some form of counterfactual bias.

These findings suggest that while LLMs outperform humans in accuracy, they are likely to reinforce existing biases when assisting in decision-making.

\paragraph{LLMs are less susceptible to framing effects}
One key result from \citep{abels2024mitigating} was that humans showed significant framing effects. In particular, humans often gave similar likelihood ratings to headlines reporting opposite outcomes for age and ethnicity categories. This was attributed to varying levels of skepticism: humans showed low skepticism for age-related headlines, assigning high likelihoods regardless of content, but were more skeptical of ethnicity-related headlines, often assigning lower likelihoods. Humans were more discerning for gender-related headlines but had lower skepticism for headlines reporting negative outcomes for men.

In contrast, we found that LLM groups tend to be less susceptible to framing effects than human groups. Supplementary Figure S4 shows LLMs are not biased in their responses to headlines reporting outcomes for ethnic groups. Similarly, while humans show significant framing effects for age headlines, we found that LLMs display very little framing effects, suggesting they have stronger opinions on these headlines than human responders. In terms of framing effects for gender headlines, LLMs aligned closely with humans. 

Note that, while the average response from LLMs shows framing effects, LLMs are not uniformly susceptible to them. Supplementary Figure S5 shows that 6 models (claude-3-5-haiku, claude-3-5-sonnet, gemini-1.5-pro, gemini-2.0-flash, gpt4, and gpt-4o) display no framing effects.

\begin{figure*}[ht]
\centering
\includegraphics[width=.9\textwidth]{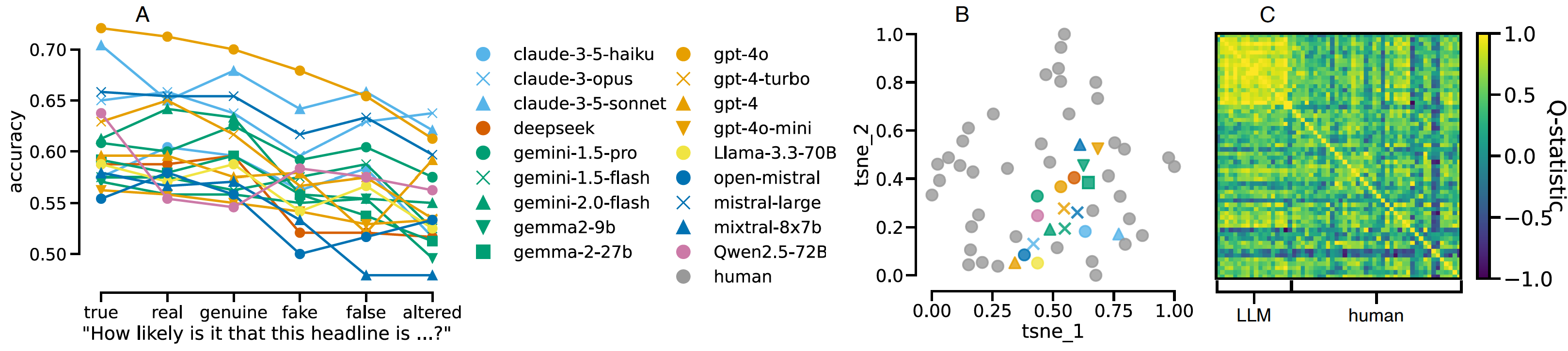}
\caption{\textbf{A.} Model accuracy across prompt variations, with each dot representing the accuracy of a model for a specific prompt ending. \textbf{B.} t-SNE visualization of responder diversity. \textbf{C.} Q-statistics (see Equation \ref{eq:qstatistic}) matrix. Each cell gives the Q-statistic between responder pairs.}
\label{fig:framing_effects_prompt}
\end{figure*}

Lastly, we investigate how different prompts affects LLM performance (\autoref{fig:framing_effects_prompt}). Results show that the original prompt ("how likely is it that this headline is true?") achieves the highest accuracy, suggesting it best captures what is being tested in the headline dataset. Variants such as ``real" and ``genuine" slightly reduce performance, while antonyms like ``fake", ``false", and ``altered" cause significant declines. This shows that while LLMs are robust against headline-induced framing effects, they remain sensitive to prompt framing.

\subsection{Wisdom and bias of LLM crowds}

Building on our understanding of individual LLM behavior, we now examine how groups of LLMs, humans, and hybrid ensembles perform collectively. 

\paragraph{Diversity of LLM crowds}
We first examine the diversity of LLM crowds through the Q-statistic (Equation \ref{eq:qstatistic}). 

\autoref{fig:framing_effects_prompt}.C displays the Q-statistic between the 18 LLMs and 40 of the human participants, while \autoref{fig:framing_effects_prompt}.B clusters responders based on their Q-statistics. %Clusters of models indicate groups with similar error patterns or predictive behaviors, suggesting underlying similarities in their learned representations. 

Both the close proximity of LLM models in \autoref{fig:framing_effects_prompt}.B and the high values in the LLM sub-matrix of \autoref{fig:framing_effects_prompt}.C show that LLMs exhibit much higher correlation among themselves than humans. Specifically, the average Q-statistic within human ensembles is $0.387  \pm 0.33$, while that of LLM ensembles is $0.855  \pm 0.08$. 
This implies that when one LLM makes a mistake, others are likely to make the same mistake. Consequently, simply averaging LLM outputs is unlikely to improve performance and may even lead to consensus on incorrect answers, amplifying shared biases.

Notably, the average Q-statistic for hybrid ensembles is $0.548 \pm 0.31$, significantly lower than that of LLM-only ensembles. This suggests that hybrid groups could complement LLM accuracy with the diversity of human responders.

\paragraph{Static aggregates reinforce biases}
The wisdom of the crowd relies on diversity within the group to cancel out individual mistakes. 
However, simple averages carry individual LLM biases into the aggregate (Table \ref{tab:performance}, average(LLM) row), as they lack the diversity---e.g., being biased in opposite directions---to allow averaging to mitigate biases. Notably, while only a minority of models show bias for the \textsc{Gender}-altered categories, their aggregate is significantly biased. 

The lack of diversity among LLMs also results in smaller gains from aggregation compared to human groups. For example, aggregation increases human performance from an individual average of $0.55$ to $0.611$. In contrast, LLM groups improve only slightly, from $0.61$ to $0.652$. Restricting LLM groups to high-performing LLMs (i.e., LLM+, see \autoref{sec:ensembling}) slightly raises the aggregated performance to $0.654$, but biases remain unmitigated.

Hybrid ensembles (average(hybrid)), achieve performance levels between purely human and LLM groups. In contrast, selective hybrid ensembles (average(hybrid+)) significantly boost performance, even outperforming groups of exclusively strong LLMs (average(LLM+)). While LLM groups offer strong individual accuracy, their high correlation limits the benefits of additional group members. Partially replacing LLMs with randomly sampled humans introduces greater diversity, allowing the collective to correct more errors and achieve stronger overall performance.

\paragraph{Group Size and Collective Intelligence}
Condorcet's Jury Theorem \citep{condorcet1785essay} and related principles suggests that larger, diverse groups should achieve higher accuracy, assuming members are reasonably independent and perform better than chance \citep{surowiecki2005wisdom}.
 
\begin{figure}[h]
\centering
\includegraphics[width=.45\textwidth]{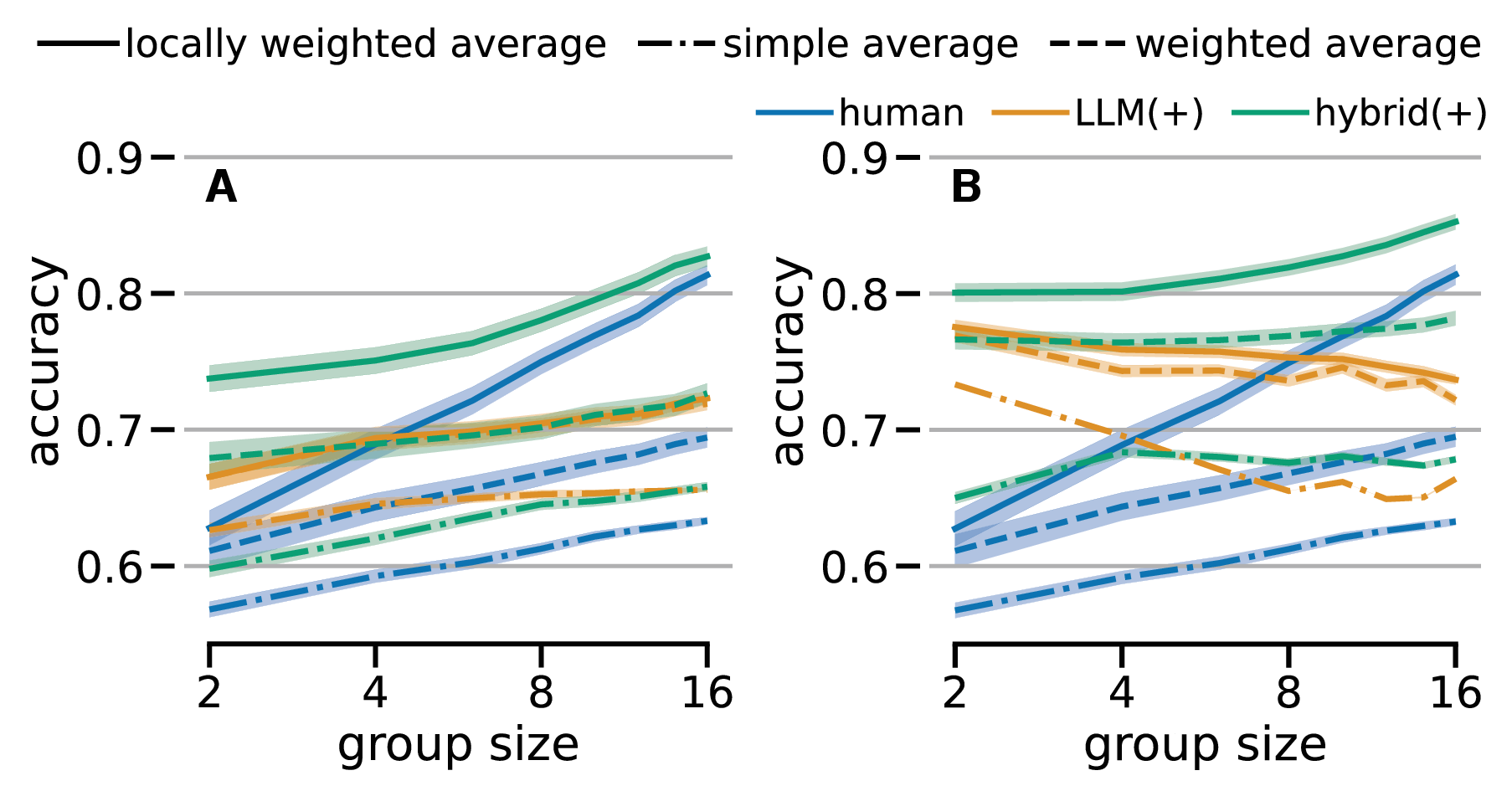}

\caption{Accuracy for different group sizes and aggregators. Shaded areas show $95\%$ confidence intervals. LLMs are either sampled randomly (\textbf{A}) or based on their MMLU scores (\textbf{B}).}
\label{fig:perf_per_group_size}
\end{figure}

Figure \ref{fig:perf_per_group_size}.A shows that, for the simple average, the limited diversity of LLM crowds results in diminishing returns as group size increases.   In contrast, human groups exhibit steady performance gains due to greater diversity. Hybrid groups initially perform between LLM and human groups but catch up to LLMs as group size increases, suggesting they benefit from both LLM accuracy and human diversity.

When LLMs are selected for inclusion based on their MMLU benchmark scores (\autoref{fig:perf_per_group_size}.B), the simple average improves significantly. However, as additional LLMs tend to be similar to, but weaker than, already included LLMs, increasing group size leads to decreased performance, as the aggregate is unable to benefit from weaker or redundant contributions. This also explains the stagnation of the simple average for hybrid ensembles beyond group size $4$; adding LLMs introduces redundancy, while adding humans provides beneficial diversity. Beyond size $4$, these effects cancel each other out, resulting in plateauing performance.

%Although LLM averages outperform human averages despite their lack of diversity, this limitation becomes more pronounced with advanced aggregation methods. These methods rely on identifying responders with uncorrelated error patterns to maximize collective performance. The high correlation among LLMs makes this challenging, though their higher individual accuracy partially compensates for the lack of diversity.

%To better understand this trade-off, we now examine (locally) weighted aggregates.

\paragraph{ExpertiseTrees promote the wisdom of the crowd}
\autoref{fig:perf_per_group_size}.A demonstrates that locally weighted averages derived from ExpertiseTrees consistently outperform simple averages. While simple LLM averages stagnate beyond group size 4, locally weighted averages continue to improve by leveraging additional LLMs. Interestingly, although small LLM groups outperform human groups, human diversity leads to higher performance in large groups.

The most significant improvements occur in hybrid groups, which outperform both human and LLM groups. To balance the high accuracy of LLMs with the diversity of humans, ExpertiseTrees assign higher weights to well-performing individuals (typically LLMs) and to complementary subgroups of humans. In addition, by potentially maintaining a distinct set of weights for each category, the ExpertiseTree can also capitalize on the specialized strengths of certain human responders. A comparison to regular weighted averages highlights the benefits of this specialization, as ExpertiseTrees consistently outperform them when the group contains humans. 

When LLMs and hybrid groups are selected based on MMLU benchmark scores (\autoref{fig:perf_per_group_size}.B), performance further improves. In particular, by having a more informed selection of LLMs, their aggregation outperforms human groups for more group sizes. Note that despite the use of ExpertiseTrees, including more LLMs still decreases performance, as without any useful diversity or improved individual accuracy, the additional LLMs simply introduce more noise. Conversely, hybrid groups continue to benefit from the inclusion of humans, as their diversity complements LLM accuracy. 

\paragraph{ExpertiseTrees mitigate biases}
Beyond simply improving accuracy, Table \ref{tab:performance} shows that ExpertiseTrees mitigate biases.  In particular, the ExpertiseTree(LLM) and ExpertiseTree(LLM+) results in fewer significant biases compared to simple and weighted LLM(+) averages. However, the most prevalent bias in individual LLMs persists in the ExpertiseTree aggregates, likely because the lack of diversity within LLM-only groups limits the capacity for bias mitigation.

In contrast, the diversity within human and hybrid crowds allows ExpertiseTrees to mitigate biases displayed by individuals and (weighted) averages of human or hybrid groups.

\section{Conclusion}\label{sec:conclusion}
In this work we investigated whether LLMs exhibit similar biases to humans on the headline dataset \citep{abels2024mitigating}. Human responses to this dataset revealed susceptibility to counterfactual biases and framing effects. Our findings confirm that LLMs reflect those counterfactual biases. All tested LLMs exhibited significant biases, with many favoring headlines reporting positive outcomes for White individuals over Black individuals. Approximately half of the LLMs showed a similar bias favoring women over men. However, LLMs were less affected by framing effects from headline variations, showing greater consistency. Nevertheless, LLMs were susceptible to prompt-induced framing effects, with subtle changes in wording significantly impacting accuracy.

To mitigate individual errors and biases, we explored the ``wisdom of the LLM crowd". Our initial experiments revealed that simple averaging of LLM outputs marginally improved accuracy but reinforced existing biases due to the lack of diversity among LLMs. In contrast, locally weighted averages partially restored the benefits of the crowd, mitigating some biases while improving performance.

Recognizing the complementary strengths of humans (greater diversity) and LLMs (higher individual accuracy), we investigated hybrid crowds that combine both. We found that hybrid crowds outperformed purely human and purely LLM groups in both simple and weighted aggregation approaches. Notably, while locally weighted averages of LLM groups still exhibited counterfactual biases, hybrid crowds achieved improved accuracy without significant biases. 

To conclude, our findings highlight the potential of integrating humans and AI within collective intelligence systems. Even modestly sized hybrid ensembles demonstrated advantages, combining the accuracy of LLMs with the diversity of human perspectives to achieve more robust and fair outcomes.

\paragraph{Limitations}
While our findings are promising, there are a few important considerations. First, the analysis was conducted using a single dataset. While this provided a controlled environment for systematically comparing biases and performance across LLMs and humans, results may differ for datasets that have different cultural or demographic contexts.

Second, the observed biases and performance reflect the specific versions of LLMs tested. As these models are frequently updated and retrained, future versions may exhibit different behaviors, potentially affecting our findings.

Third, our exploration of diversity focused on hybrid ensembles but did not incorporate techniques to engineer diversity within LLMs, such as fine-tuning or prompting models to adopt varied personas or perspectives. Such approaches could further enhance ensemble diversity and offer additional opportunities for bias mitigation.

\section*{Acknowledgments}
The author(s) disclosed receipt of the following financial support for the research, authorship, and/or publication of this article: A.A. is supported by a post-doctoral grant (Chargé de Recherche) by the National Fund for Scientific Research (F.N.R.S.) of Belgium. T.L. is supported by the F.N.R.S. [grant numbers 31257234 and 40007793], the Fonds Wetenschappelijk Onderzoek (F.W.O.) [grant number G.0391.13N], the Service Public de Wallonie Recherche [grant n\textdegree 2010235–ARIAC by DigitalWallonia4.ai], the Flemish Government through the AI Research Program, and TAILOR, a project funded by EU Horizon 2020
research and innovation program [grant number 952215]. The resources and services used in this work were provided by the VSC (Flemish Supercomputer Center), funded by the Research Foundation - Flanders (FWO) and the Flemish Government.

\appendix

\bibliographystyle{named}
\bibliography{ijcai25}

\newpage 
\makeatletter
\renewcommand \thesection{S\@arabic\c@section}
\renewcommand\thetable{S\@arabic\c@table}
\renewcommand \thefigure{S\@arabic\c@figure}
\makeatother

\twocolumn[
\begin{center}
    {\Large Supplementary Information for 
    
    \textit{Wisdom from Diversity: Bias Mitigation Through Hybrid Human-LLM Crowds}}
\end{center}

\vspace{1em}
]
\section{Supplementary Information}
\maketitle
\begin{table*}[h!]
\centering
\begin{tabular}{lllcll}
\toprule
\textbf{Name} & \textbf{Version} & \textbf{Size} & \textbf{MMLU} & \textbf{Company} & \textbf{Provider} \\
\midrule
Qwen2.5-72B-Instruct  &  & 72B & 86.1 & Alibaba & HuggingFace \\
claude-3-5-haiku  & 20241022 & 20B$^1$ & 75.2 & Anthropic & Anthropic\\
claude-3-5-sonnet  & 20240229 & 137B$^1$ & 88.7 & Anthropic & Anthropic\\
claude-3-opus  & 20240229 & 137B$^1$ & 86.8 & Anthropic & Anthropic\\
DeepSeek & V3 & 18x37B & 87.1 & DeepSeek AI & DeepSeek AI\\
gemini-1.5-flash & September 2024 & 32B$^1$ & 78.9 & Google & Google\\
gemini-1.5-pro & September 2024 & 120B, 1.5T$^1$ & 85.9 & Google & Google \\
gemini-2.0-flash-exp & December 2024 & 32B$^1$ & 87 & Google & Google \\
gemma-2-9B &  & 9B & 71.3 & Google & Groq\\
gemma-2-27B &  & 27B & 75.2 & Google & HuggingFace\\
gpt-4 & 0613 & 175B$^1$ & 86.4 & OpenAI & OpenAI\\
gpt-4-turbo & 2024-08-06 & 175B$^1$ & 86.4 & OpenAI & OpenAI\\
gpt-4o & 2024-08-06 & 1T$^1$ & 88.7 & OpenAI & OpenAI\\
gpt-4o-mini & 2024-07-18 & 8B$^1$ & 82 & OpenAI & OpenAI\\
Llama3.3-70B  &  & 70B & 82 & Meta & HuggingFace\\
mistral-large-latest  & 24.11 & 123B & 84 & Mistral & Mistral\\
mixtral-8x7b  &  & 8x7B & 70.6 & Mistral & Groq\\
open-mistral-nemo  &  & 12B & 68 & Mistral & Mistral\\
\bottomrule
\end{tabular}
\caption{Model summary. $^1$ denotes estimates from various sources of closed models based on their performance, response time, and cost. MMLU column provides the reported performance of the models on the MMLU benchmark \citep{hendrycks2020measuring}. }
\label{tab:model_comparison}
\end{table*}

\autoref{tab:model_comparison} presents an overview of the LLMs we considered in this work, including the version, estimates of the number of parameters, their MMLU \citep{hendrycks2020measuring} score, the company which initially developed them, as well as the provider we used. Note that several models are proprietary, and their sizes are therefore estimated based on their cost, their response times, as well as their performance. 

\subsection{Prompt}
\autoref{fig:prompt} shows the prompt we presented to the LLMs:
\begin{figure*}
\begin{verbatim}
How likely is it that this headline is {target_str}.

 Choose one of the following options and return only the number of that option:
1. very unlikely, 2. unlikely, 3. undecided, 4. likely, 5. very likely.

<examples>
"{example_headline1}"
Response: {expected_response1}
"{example_headline2}"
Response: {expected_response2}
"{example_headline3}"
Response: {expected_response3}
"{example_headline4}"
Response: {expected_response4}
</examples>

"{response_headline}"
Response: 
\end{verbatim}
\caption{The prompt presented to the LLMs. }\label{fig:prompt}
\end{figure*}

Where, like in the prompt presented to human participants in  \citep{abels2024mitigating}, \verb|target_str| is ``true" unless specified otherwise. For \autoref{fig:framing_effects_prompt} we explore different values for \verb|target_str|. 

As multi-shot learning \citep{brown2020language} has been shown to improve accuracy and compliance with instructions (here, the expected answer format), we include for each headline 4 example headline-response sequences. Example headlines are sampled randomly from the same category (i.e., age headlines will be presented with examples on age headlines) and are balanced across the different statuses (genuine or altered) and sentiments (positive or negative). Since the dataset contains both genuine and altered variants of the same headline, we ensured that the examples never contained the alternative version of the queried headline. 
To reduce variability across models, we fixed the temperature to $0$, such that the most likely token was returned. 
The responses are then parsed to extract the label ([1-5]), which, as in \citep{abels2024mitigating}, are then mapped to numeric values $\{0,0.25,0.5,0.75,1\}$.  

Responses which failed to answer the query, such as those containing system-level disclaimers (“As an AI model, I…”) were re-queried for consistency.

 \subsection{ExpertiseTrees}

ExpertiseTrees \citep{pmlr-v202-abels23a} are an advanced aggregation method designed to partition the context space (e.g., headline categories) and fit localized aggregations for more effective decision-making. They extend the principles of decision trees by incorporating model-based predictions at the leaves, enabling them to dynamically adapt to varying contexts while maintaining interpretability.

\subsubsection{Structure and Function}

Similar to decision trees, ExpertiseTrees recursively split the problem space, with each split chosen to maximize the performance improvement of the tree. Nodes in the tree correspond to subsets of the context space, such as gender-related or ethnicity-related headlines. This partitioning enables ExpertiseTrees to isolate specific contexts where distinct prediction patterns or biases may occur, tailoring the aggregation process to the nuances of each subset.

Unlike traditional decision trees, where the leaves contain constant values or simple averages, the leaves of an ExpertiseTree contain models. Specifically, these models are linear combinations of individual predictions, weighted to reflect the relevance and reliability of each member's contribution within that context. This allows ExpertiseTrees to dynamically adapt aggregation weights based on context-specific patterns, enhancing both accuracy and fairness.

\subsubsection{Advantages Over Traditional Approaches}

ExpertiseTrees offer several advantages compared to static aggregation methods like simple averaging or traditional stacking \citep{breiman1996stacked}:

\begin{itemize}
    \item \textbf{Context Sensitivity}: By partitioning the context space, ExpertiseTrees can identify and exploit differences in performance or bias across contexts (e.g., headline categories). For example, an ExpertiseTree might assign higher weights to certain predictors for gender-related headlines and adjust those weights for ethnicity-related headlines.
    \item \textbf{Dynamic Weighting}: Unlike static methods that assign fixed weights to predictors, ExpertiseTrees adjust weights at the leaf level based on the specific context. This ensures that the most reliable and unbiased predictors contribute more heavily to the aggregate output within each subset of the context space.
    \item \textbf{Improved Bias Mitigation}: By isolating subsets of the data where certain individuals or models perform better or exhibit less bias, ExpertiseTrees can mitigate biases more effectively than methods that aggregate across the entire dataset. For example, if certain predictors are prone to counterfactual bias in ethnicity-related headlines but perform well in gender-related contexts, ExpertiseTrees can selectively downweight their contributions.
    \item \textbf{Interpretability}: The linear combination models at the leaves ensure that the aggregation remains interpretable, providing clear insight into how individual predictions are combined within each context.
\end{itemize}

By routing predictions through the tree structure, ExpertiseTrees effectively balance the strengths and weaknesses of individual contributors, resulting in improved accuracy and fairness across headline categories.

\subsection{Framing Effects}

\begin{figure*}[h!]
\centering
\includegraphics[width=1\textwidth]{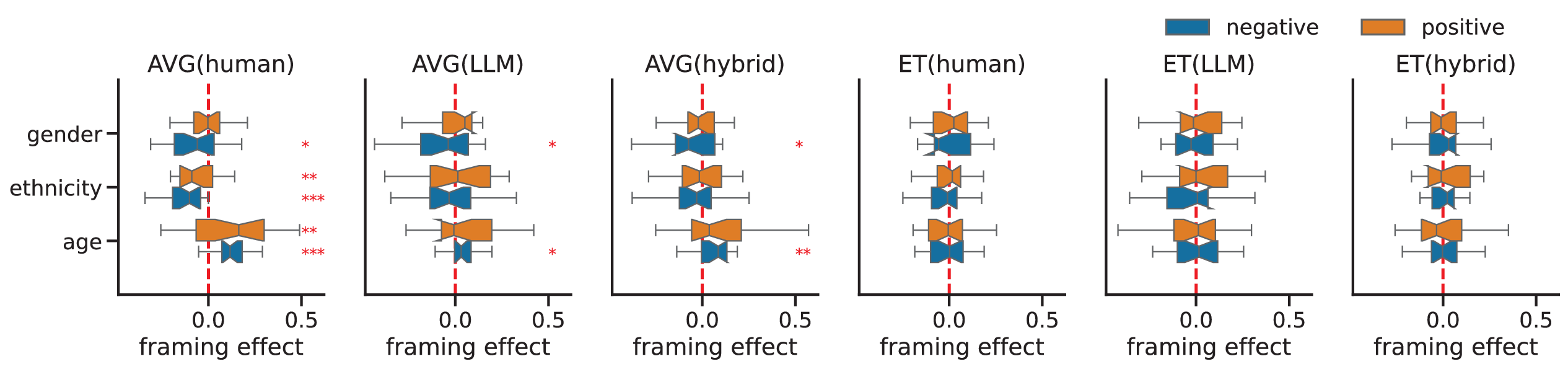}
\caption{Distribution of framing effects for headlines reporting positive or negative outcomes across three demographic groups. Asterisks indicate statistical significance of the framing effects (Wilcoxon test, *: $p<0.10$, **: $p<0.05$, ***: $p<0.01$).}
\label{fig:framing_effects}
\end{figure*}

\autoref{fig:framing_effects} visualizes the framing effects observed across the three group types (LLM-only, human-only, and hybrid) under either human or ExpertiseTree aggregation. Each boxplot represents the distribution of framing effects (see \autoref{sec:metrics}), quantified as $\Delta_F(\text{sentiment, man})$ for gender, $\Delta_F(\text{sentiment, old})$ for age, and $\Delta_F(\text{sentiment, white})$ for ethnicity. Since every headline presents a contrasting statement for a complementary demographic group, the distributions for the complementary groups (not shown here) would simply be inverted versions of those shown in \autoref{fig:framing_effects}.

\begin{figure*}[h]
\centering
\includegraphics[width=1\textwidth]{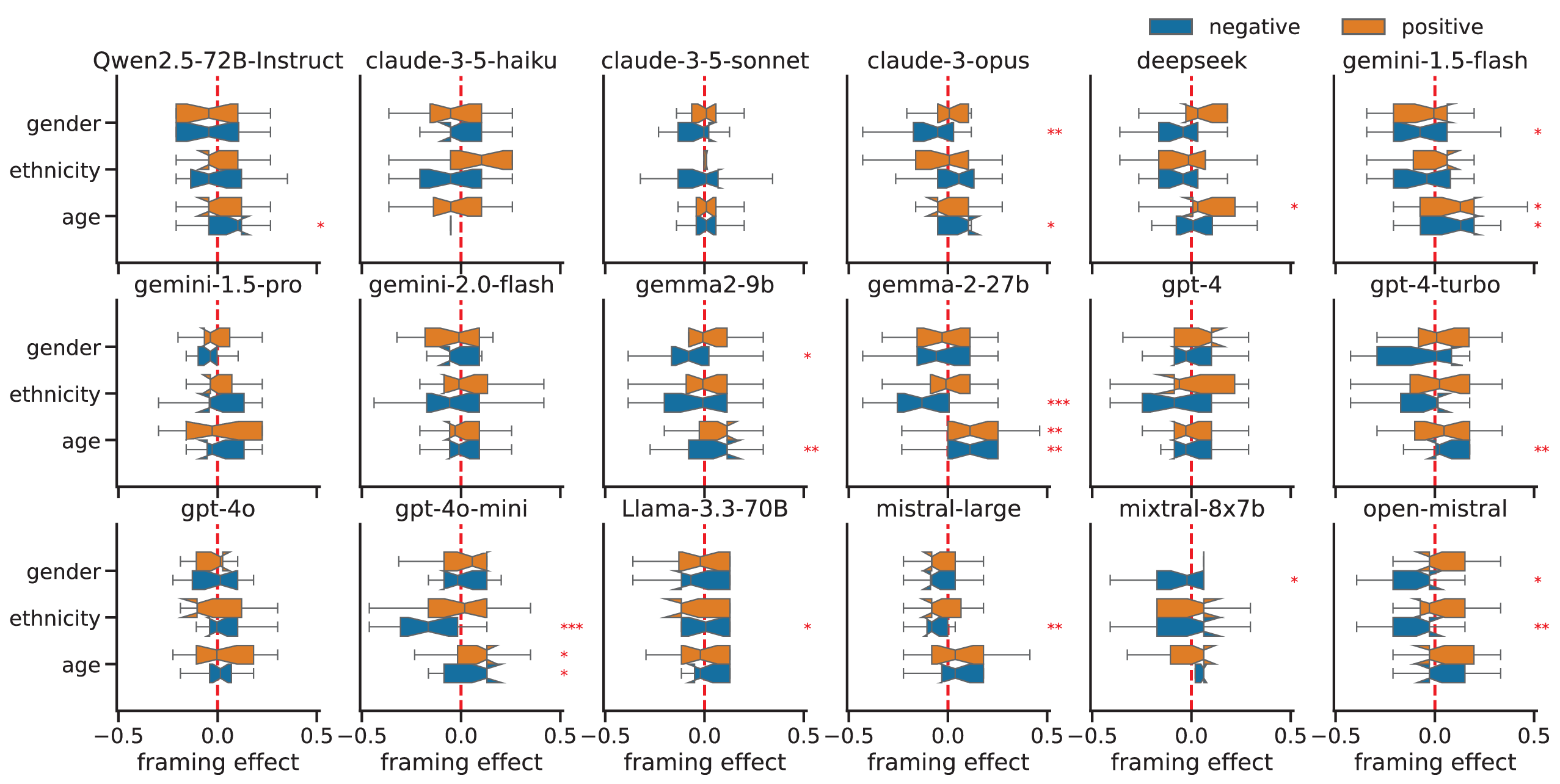}
\caption{Distribution of framing effects for headlines reporting positive or negative outcomes across three demographic groups. Asterisks indicate statistical significance of the framing effects (Wilcoxon test, *: $p<0.10$, **: $p<0.05$, ***: $p<0.01$).}
\label{fig:model_framing_effects}
\end{figure*}

\end{document}